\pdfoutput=1

\documentclass[11pt]{article}

\usepackage[]{EMNLP2022}
\usepackage[algo2e]{algorithm2e}
\usepackage{algorithm}
\usepackage{import}
\usepackage{times}
\usepackage{latexsym}
\usepackage{graphicx}
\usepackage{multirow}
\usepackage[T1]{fontenc}

\usepackage[utf8]{inputenc}

\usepackage{microtype}

\usepackage{inconsolata}

\title{AX-MABSA: A Framework for Extremely Weakly Supervised Multi-label Aspect Based Sentiment Analysis}

\author{Sabyasachi Kamila$^{1}$, Walid Magdy$^{1}$, Sourav Dutta$^{2}$ \and MingXue Wang$^{2}$\\
  $^{1}$School of Informatics, University of Edinburgh \\
$^{2}$Huawei Research Centre, Dublin, Ireland  \\
  \texttt{\{skamila,wmagdy\}@inf.ed.ac.uk},
  \texttt{\{sourav.dutta2,wangmingxue1\}@huawei.com} \\}

\begin{document}
\maketitle
\begin{abstract}
Aspect Based Sentiment Analysis is a dominant research area with potential applications in social media analytics, business, finance, and health. Prior works in this area are primarily based on supervised methods, with a few techniques using weak supervision limited to predicting a single aspect category per review sentence. In this paper, we present an extremely weakly supervised multi-label {\em Aspect Category Sentiment Analysis} framework which does not use any labelled data. We only rely on a single word per class as an initial indicative information. We further propose an automatic word selection technique to choose these seed categories and sentiment words. We explore unsupervised language model post-training to improve the overall performance, and propose a multi-label generator model to generate multiple aspect category-sentiment pairs per review sentence. Experiments conducted on four benchmark datasets showcase our method to outperform other weakly supervised baselines by a significant margin.\footnote{Code, data and model are available at \url{https://github.com/sabyasachi-kamila/AX-MABSA}} 
\end{abstract}

\section{Introduction}

Aspect-based sentiment analysis (ABSA) is a well-known sentiment analysis task which provides more fine-grained information than simple sentiment understanding \cite{liu2012sentiment}. The main goal of ABSA is to find the aspects and its associated sentiment within a given text. While the works on ABSA have expanded in different directions, it has primarily two sub-tasks, {\em Aspect Term Sentiment Analysis} (ATSA) and {\em Aspect Category Sentiment Analysis} (ACSA) \cite{xue2018aspect}. ATSA consists of different tasks like aspect term extraction \cite{li2018aspect, luo2019unsupervised, li2020conditional, shi2021simple}, aspect term sentiment classification \cite{he2018exploiting, chen2019transfer, hou2021graph}, opinion term extraction \cite{dai2019neural, he2019interactive, chen2020relation}, aspect-oriented opinion term extraction \cite{fan2019target, wu2020grid}, aspect-opinion pair extraction \cite{zhao2020spanmlt}, etc. For example, in the sentence ``\textit{The sushi is top-notch, the waiter is attentive, but the atmosphere is dull.}", ATSA would extract the aspect terms `\textit{sushi}', `\textit{waiter}' and `\textit{atmosphere}'; opinion terms `\textit{top-notch}', `\textit{attentive}', and `\textit{dull}'; and their associated sentiments `\textit{positive}', `\textit{positive}' and `\textit{negative}'. The other sub-task ACSA aims to find the higher order aspect categories and its associated sentiment from a given text. In the above example, ACSA would detect the categories as `\textit{food}' (as `pasta' is a type of `food'), `\textit{service}' and `\textit{ambience}'; and the associated sentiments as `\textit{positive}', `\textit{positive}' and `\textit{negative}'. 

Existing research on ABSA is dominated by supervised methods, where labeled training data is provided \cite{chen2017recurrent, xue2018aspect, cai2021aspect, liu2021solving, xu2021learning, yan2021unified}. A few works try to solve the problem in a weakly/semi-supervised manner, where a few labelled samples are provided \cite{wang2021progressive}. However, there has been a lack of study on ABSA using {\em unsupervised methods}, i.e., without using any labelled data. A few works also focused on unsupervised aspect term extraction \cite{shi2021simple}. However, such works do not deal with the sentiment associated with the aspects. An existing work on weakly supervised ACSA \cite{huang2020weakly} only considered a single aspect category per sentence -- thus limiting the task to a larger extent. 

Motivated by the above, in this work, we present a methodology for {\em extremely weakly supervised ACSA} task, where we do not need any labelled training samples. We solve both aspect category detection (ACD) and ACSA tasks (on each review sentence) just by using the surface text of aspect category and sentiment. Given \textit{N} review sentences, \textit{C} categories of interest and \textit{P} polarities of interest, the ACD task generates \textit{C} clusters, while the ACSA task generates \textit{($c_i$, $p_j$)} tuples where $\textit{$c_i$} \in \textit{C}$, and $\textit{$p_j$} \in \textit{P}$. As in \cite{wang2021x}, we adopt the representation learning perspective, wherein representing sentences by class names leads to better clustering. We only use the surface text of the class names and unlabelled sentences to get aspect category and sentiment clusters.

However, in clustering, each review sentence would get only one label, thus limiting the task by a substantial extent. To tackle this, we propose {\em X-MABSA}, a {\em multi-label generator model} which makes use of dependency parser \cite{qi2020stanza} and a similarity-based attention mechanism to generate multiple categories and associated sentiment polarity labels for each review sentence. In addition, we find that sometimes the representative text of aspect categories (provided as input) is not present (or sparse) in the text corpus. This might lead to skewed representation of the classes in our framework and thus degrade performance. Therefore, we present an automatic surface word selection strategy which would represent the class names better. We combine this with our X-MABSA model and denote it as AX-MABSA. 

We also showcase that unsupervised post-training of language model on domain specific data significantly improves the sentence representation and thus achieves better results for ACSA tasks. For this, we post-train BERT language model \cite{devlin2019bert} using domain specific unlabelled data. We perform experiments on four different benchmark aspect-based datasets \cite{pontiki-etal-2014-semeval, pontiki-etal-2015-semeval, pontiki-etal-2016-semeval, cheng2017aspect}, and compare with different supervised and weakly supervised baselines.

Our main contributions are as follows: 

\begin{itemize}

\item an extremely weakly supervised method to solve the ACSA task without relying on any labelled data, and using only the class names as the only provided information; 

\item an automatic surface word selection strategy for choosing a suitable word corresponding to each aspect and sentiment class; 

\item use of BERT language model post-training on domain specific unlabelled data for semantic representation of review sentences; 

\item a multi-label generator model which makes use of a dependency parser and a similarity-based attention mechanism for generating multiple aspect-sentiment labels for each sentence; and 

\item experimental results comparing our architecture with different existing baselines on four benchmark aspect datasets.
\end{itemize}

\section{Related Work}
Aspect Based Sentiment Analysis (ABSA) has gained significant attention for a long time, and research has been done in primarily two directions -- Aspect Term Sentiment Analysis (ATSA) and Aspect Category Sentiment Analysis (ACSA).

\subsection{Aspect Term Sentiment Analysis} 
Research on ATSA has been in different sub-categories like, 
\paragraph{Aspect Term Extraction} 
In this sub-task, aspect terms associated with a category are extracted from a given text. Prior research on this is based on sequence labelling problem \cite{ma2019exploring, li2020conditional}. \citet{li2017deep} proposed a neural network-based deep multi-task framework with memory network for extracting aspect terms. \citet{xu2018double} presented a double embedding method which uses CNN \cite{lecun1995convolutional}-based sequence tagging, while \citet{li2018aspect} considered summary of opinions expressed in text as well as the history of aspect detection for effective aspect term extraction. \citet{chen2020enhancing} proposed a soft prototype-based approach with aspect word correlations to improve quality. A few unsupervised methods have tried to improve performance by using traditional topic modelling-based models. \citet{luo2019unsupervised} proposed a neural network based unsupervised model which takes sememes for better lexical semantics. \citet{shi2021simple} presented a self-supervised method which works on learning aspect embedding on the word embedding space for aspect extraction. 

\paragraph{Aspect-level Sentiment Classification} 
In this sub-task, sentiment labels are assigned to each aspect term. \citet{wang2016attention, liu2017attention, ma2017interactive} proposed an attention-based neural network model for aspect-level sentiment classification (ASC). \citet{tay2018learning} modelled relationship between words and aspects using LSTM model \cite{hochreiter1997long} to improve ASC performance. \citet{he2018exploiting} showed that document knowledge transfer improved performance of ASC task. \citet{chen2019transfer} proposed a transfer capsule network for transferring knowledge from document-level sentiment classification, while \citet{hou2021graph} adopted a dependency tree-based graph neural network to solve the ASC task. 

\paragraph{Aspect-oriented Opinion Extraction} 
This task extracts opinion terms associated with aspect terms. \citet{fan2019target} designed a sequence label model which used LSTM \cite{hochreiter1997long} for aspect-oriented opinion extraction (AOE). \citet{wu2020grid} proposed a tagging scheme for AOE task which uses CNN \cite{lecun1995convolutional}, LSTM \cite{hochreiter1997long} and BERT \cite{devlin2019bert} for opinion extraction. \citet{wu2020latent} proposed a transfer learning method for transferring knowledge from sentiment classification task to AOE task.

Recent works on ATSA have introduced more sub-tasks like aspect-opinion pair extraction, aspect-sentiment-opinion triplet extraction, aspect-category-opinion-sentiment quadruple extraction, etc. \citet{yan2021unified} proposed a BART \cite{lewis2020bart} -based model to solve all ATSA tasks. \citet{cai2021aspect} introduced a new task called, aspect-category-opinion-sentiment quadruple extraction, a BERT \cite{devlin2019bert}-based model to deal with implicit aspects and opinion terms. \citet{xu2021learning} proposed a new span-level method for the aspect-sentiment-opinion triplet extraction.

\subsection{Aspect Category Sentiment Analysis} 
Aspect Category Sentiment Analysis (ACSA) finds aspect categories and their associated sentiments from a text. Research on this has been conducted on both Aspect Category Detection (ACD) and ACSA tasks. \citet{ma2018targeted} proposed a word attention-based hierarchical model which takes common-sense knowledge for solving ACSA task. \citet{xue2018aspect} presented a novel CNN \cite{lecun1995convolutional}-based model for ACSA task. \citet{liang2019novel} proposed an encoding scheme which was aspect-guided and able to perform aspect-reconstruction. \citet{sun2019utilizing} constructed an auxiliary text for aspects and reformed the ACSA as a classification task.

\citet{wang2020relational} proposed a novel dependency tree-based model and a relational graph attention network for encoding the sentences. \citet{li2020multi} designed a multi-instance framework for multi-label ACSA task. \citet{cai2020aspect} reformed the task as sentiment-category with a two-layer hierarchy where the higher layer detected the sentiment while the lower layer detected the aspect category. \citet{liang2021beta} presented a semi-supervised framework having a beta distribution-based model. The model finds semantically related words from the context of a target aspect. \citet{liu2021solving} solved the ACSA task as a text generative method using BART \cite{lewis2020bart}. \citet{zhang2021aspect} presented aspect sentiment quad prediction task where ACSA was formulated as a paraphrase generation task.

Almost all existing works on ACSA are based on supervised methods. In contrast, this work proposes a method for ACSA which does not require any labelled data and relies only on seed text for aspect class names. 
\section{Proposed Methodology}
\begin{figure*}
    \centering
    \includegraphics[width=0.9\textwidth]{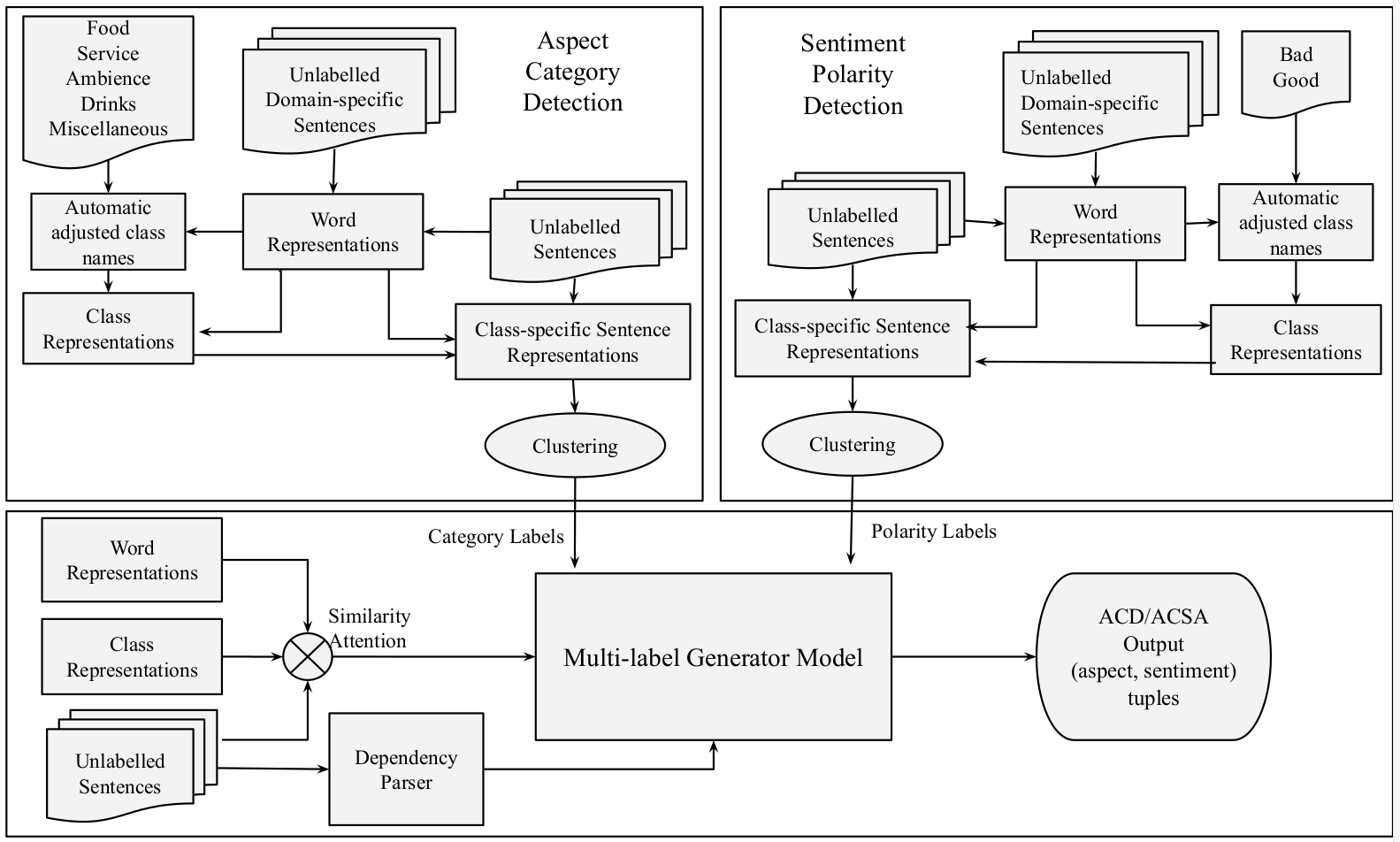}
    \caption{Overview of the proposed AX-MABSA Framework}
    \label{fig:1}
\end{figure*}

Our proposed method, {\em AX-MABSA}, works on the following components: (a) class name-based clustering, (b) unsupervised language model post-training on domain-specific data for better contextual representation of review sentences, (c) a multi-label generator model to generate multiple categories and associated sentiment labels, and (d) automatic class-representative text selection. The overall framework is depicted in Figure \ref{fig:1}.

\paragraph{\textbf{Problem Formulation:}} 
We formulate the extremely weakly supervised ACD and ACSA tasks as: Consider as input a review sentence $x= \{x_{1}, x_{2}, x_{3}, ......, x_{n}\}$ where $x_{i}$ is the $i^{th}$ word of the sentence and $n$ is the length of the sentence, along with a list of \textit{C} predefined aspect categories. The output for the ACD task is $c$ categories for a sentence where $c\subset C$. For the ACSA task, the output is a list of tuples ($c_{j}$, $p_{k}$) where $c_{j}$ is the $j^{th}$ predicted category and $p_{k}$ is the $k^{th}$ predicted sentiment polarity corresponding to the category $c_{j}$. The sentiment polarity p is from the set $s \subset \{positive, negative\}$.

\subsection{ACSA Module}
As a primary task, we address the aspect detection based on the seed aspect categories provided as input. We adopt the X-Class model as presented in \cite{wang2021x} for solving extremely weakly supervised classification tasks majorly on topic modelling datasets. This module involves four stages: (a) word representations, (b) class representation, (c) class-specific document representation, and (d) document-class alignment.  

For word representations, at first, a vocabulary is created from all the input texts. Then, each word's contextual representation is represented using a pre-trained BERT language model \cite{devlin2019bert}. The contextual embeddings of each word are averaged to obtain the review sentence encoding, and this representation is denoted as the static word representation $s_{r}$.
\begin{equation}
    {s_{r}} = \frac{\sum_{R_{i,j}=r}{z_{i,j}}}{\sum_{R_{i,j}=r}{1}}
\end{equation}
Here, $z_{i,j}$ is the contextualized representation and $R_{i,j}$ is the $j^{th}$ word in the review sentence $R_{i}$. 

As class representation, the representations of the aspect class names are constructed based on the static representations of those words. For example, the category ``sports'' is represented by the contextual embedding of the word ``sports''. Then an expansion technique is used to find similar words of each class name words from within the input texts and average those words' contextual representations to obtain the final aspect class embedding.

In class-specific document representation, the representations of the documents or the sentences are guided by the class representations so that the sentences become more aligned to the topics of interest, i.e., the class names. Different attention mechanisms are used over the document representations guided by class representations to get updated document representations. 
Finally, for document-class alignment, clustering algorithms are used to cluster n-documents to c-clusters ($c$ is the number of classes), wherein the seed class centroids are initialized with the class representations.  

\paragraph{Clustering Algorithms:}
We followed different centroid-based clustering algorithms such as K-Means \cite{lloyd1982least}, Mini-batch K-Means \cite{sculley2010web} and Gaussian Mixture Model (GMM) \cite{duda1973pattern}; and found that in-general Mini-batch K-Means (mk-means) performs best for the ACD task while GMM performs best for the ACSA task. So, we fix this for our experiments. We used Principal component analysis \cite{abdi2010principal} for dimensionality reduction of sentence representation and class representation vectors before clustering. The target dimension is set to 64. We also fixed \textit{random\_state} to 42 for centroid initialization. For mk-means, we used batch size 400.

The model requires the surface text of the class names to be present on the dataset for a certain number of times. We feel this is a potential drawback in solving our ACSA task, as some surface text of category names may not be present in the dataset or have a proper meaning representation. For example, the category word ``miscellaneous'' might have no clear meaning and sometimes might not be present in the dataset. To resolve this issue, we explicitly add the category name to the vocabulary set if it is not found on the dataset. Another drawback of the above approach is that it can only predict one label per sentence. This is a huge limitation, especially when multiple aspect categories are present in a review sentence. In the following sections, we tackle these issues to propose a robust multi-aspect extraction framework.

\subsection{AX-SABSA}
\label{x-sabsa}
We observed that the performance of the implemented ACSA module based on X-Class is poor. One of the reasons is that the words' representations are based on the pre-trained BERT language model \cite{devlin2019bert} which gives more general representations of each word, which works well for topic modelling tasks. However, the aspect terms are more specific to the domains and thus general representation does not provide specific information. Therefore, we suggest that unsupervised post-training of BERT on domain-specific data would lead to better word-representations and thus a better performance.

\paragraph{Unsupervised Post-training of BERT Language Model (UPBERT):}
We follow a recent model \cite{gao2021simcse} which feeds the same input twice, one with dropout masks and the other with different dropout masks, to the encoder. The model optimizes the following objective function: 
\begin{equation}
\small
    z_{i}=-log\frac{e^{sim(a_{i}^{p_{i}}, a_{i}^{p_{i}^{'}})/\gamma }}{\sum_{j=1}^{N}e^{sim(a_{i}^{p_{i}}, a_{j}^{p_{j}^{'}})/\gamma }}
\end{equation}
Here, $a$ is hidden state, which is a function of the input sentence and dropout masks p and $p^{'}$.

We feed our collected domain specific unlabeled data of varied sizes to this representation and fine-tune the BERT model. For our experiments, we use \textit{batch size} as 128, \textit{sequence length} as 32, \textit{learning rate} as 3e-5, loss function as \textit{Multiple Negatives Ranking Loss} of the sentence-transformer model \cite{reimers2019sentence}. We vary the dataset size starting from 10k samples for training, and get different fine-tuned BERT models corresponding to different data sizes. Finally, we select the fine-tuned model which provides the best performance (using 80k unlabeled training sentences in our case) and apply this UPBERT model for word representation to solve the ACSA task and call our single-label predictor model X-SABSA.

\paragraph{Automatic Class-representative Surface Text Selection Algorithm (ACSSA):}

Our model suffers when the surface text of class names is not present on the data. Although we add these words to the vocabulary explicitly, their contextual representations become poor. As an immediate solution, we can manually select candidate words corresponding to class names. However, this would be a difficult and tedious job when the number of categories would be high. Also, there can be multiple candidate words for a class name. For example, to represent the category ``ambience'', one can choose any of the following words: atmosphere, environment, vibes, etc. Similarly, to represent a negative polarity, one can choose any of the following words: bad, problem, pathetic, poor, etc. Depending upon the words we choose, the overall performances vary significantly. So, we propose an algorithm that selects these candidate words automatically given the original class names (see Algorithm \ref{alg:1}). 

\begin{algorithm}[t]
\small
\caption{Algorithm for Automatic Class-representative Surface Text Selection}\label{alg:1}
\textbf{Input:} \textit{X} (noun for ACD, adjective for ACSA), dataset \textit{D}, vocabulary \textit{V}, class names \textit{C}

\textbf{Output:} A List \textit{selected} containing Candidate words for each class

Initialize a global array uV[], \textit{T}, targetL[], sourceL[], interL[], goalL[], selected[]

 \For{w in V}{
  $pos \leftarrow  getPosTag(w)$
  
  \If{pos == 'X'}{
   $uV \leftarrow append(w)$
   }
 }
  \For{i in C}{
  $sims \leftarrow findSimilarity(i, uV)$
  
  $sortedSim \leftarrow argSort(sims)$
  
  \For{j in C}{
   \If{i!=j}{
   $t \leftarrow Top T similar words from sortedSim[j]$
   
   $targetL \leftarrow t$
   }
   }
   $s \leftarrow Top T similar words from sortedSim[i]$
   
   $sourceL \leftarrow s$
   \For{w in sourceL}{
   \If{w not in targetL}{
   $interL \leftarrow append(w)$
   
   }
   }
   $goalL \leftarrow interL$
   
   $goalL \leftarrow ArgSortOccurance(goalL)$
   
   $selected \leftarrow firstValue(goaL)$
 }
\end{algorithm}

The algorithm ACSSA takes a particular part-of-speech tag (noun for the ACD task, adjective for the ACSA task), dataset, vocabulary and the class names as input and produces a candidate word list as output. Initially, it creates a list \textit{uV} of words from the vocabulary which has a desired part-of-speech tag. It then finds all the similar words for each class name from the list \textit{uV}. We then select \textit{top-T} values from each list. This can be varied depending upon inspection. We fixed it to 10 based on experimental results. Then the similar words for each class are sorted according to the cosine similarity scores.

Finally, we sort each list according to their number of occurrences in the dataset. We then select the topmost occurring word from each list as the aspect class representative. This would produce a single candidate word for each class. Thus, the AX-SABSA module uses ACSSA in combination with X-SABSA, to automatically generate better aspect category names.

\subsection{AX-MABSA}
Since clustering produces only one label for each review sentence, we propose a \textit{Multi-label Generator} model based on dependency parser \cite{qi2020stanza} and a similarity-based attention mechanism.  

\paragraph{Multi-label Generator Model:} 

This model takes the unlabelled sentences, the sentence representations, the category class representations, and the clustering outputs to generate multiple categories and associated sentiment polarity for each sentence. We illustrate the model using the following example: ``The food was good, but it's not worth the wait or the lousy service". The sentence has tags `(food, positive)' and `(service, negative)'. 

\begin{figure*}
    \centering
    \includegraphics[width=0.85\textwidth]{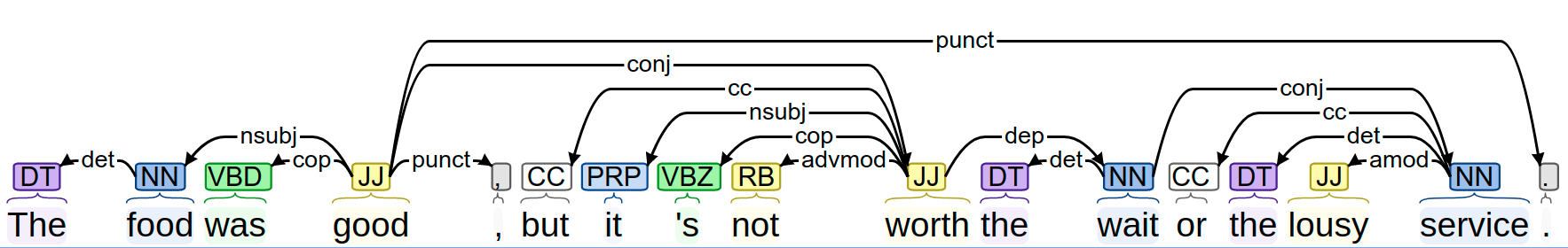}
    \caption{Sample Dependency Parser Output}
    \label{fig:2}
\end{figure*}

\textbf{Parsing the Input Sentences}
The unlabelled input sentences are parsed by off-the-shelf dependency parser \cite{qi2020stanza}. The parser outputs a pair of dependencies (word, word[head-1]). The output of the above sentence can be seen in Figure \ref{fig:2}. For each word with Noun part-of-speech tag in the sentence, we select those pairs where either the word or word[head-1] is also a Noun. We call these final set of pairs as `PPairs'. The `PPairs' for the above sentence are (`food', `good'), (`wait', `worth'), and (`service', 'wait'). Observe, in general, the first word in a pair is related to aspects while the second word is associated with sentiment. 

\textbf{Similarity-based Attention}
We use a similarity-based attention mechanism to assign a desired class label to each of the words in the PPairs. We first obtain the similarity values between the words in the sentence and all the class names using the cosine similarity as: $S_{i,j} = cos(w_i, c_j)$.

Now, we calculate $max_c(S)$ which assigns each word to the highest similar class. For each aspect word in the `PPairs' if the corresponding $max_c(S)$ is greater than a threshold\footnote{We fixed this threshold to 0.45 based on qualitative and quantitative evaluation.} then we keep those `PPairs'. We call these filtered pairs as `FPPairs'. Finally, we assign the aspect, sentiment label to each `FPPairs' based on its corresponding $max_c(S)$ values. If the `FPPairs' has only one or empty pair, then we consider the clustering outputs as the predicted aspect and sentiment pair.

In the entire setup, we use the UPBERT model mentioned in Section \ref{x-sabsa} for word representation. We refer to the entire model as X-MABSA. When we use the automatic surface word selection algorithm ACSSA in the X-MABSA model, we call that final model AX-MABSA.

\section{Experimental Setup}

We discuss here the datasets we have used, word representations, and different baselines we have selected for our experiments.\footnote{We performed our experiments on Nvidia 1e30 GPUs with CUDA 11. The post-training experiments on an average took 10-15 minutes each, and each model of our proposed method took around 1-2 minutes to run.}

\begin{table}[t]
\centering
\resizebox{0.48\textwidth}{!}{%
\small
\begin{tabular}{l|cccc}
\hline
\textbf{Dataset} & \textbf{Rest-14} & \textbf{Rest-15} & \textbf{Rest-16} & \textbf{MAMS}   \\
\hline
\# of Categories & 5 & 5 & 5  & 8  \\
\# of Sentences & 800 & 582 & 586 & 400   \\
Avg \# of Aspects/sentence &1.28 &1.21 &1.18 & 2.25\\
Imbalance & 5.04 & 12.10 & 7.34  & 9.09  \\
\hline
\end{tabular}
}
\caption{Gold Data Statistics. Imbalance value signifies the ratio between the largest and smallest category size.}
\label{table-1}
\end{table}
\subsection{Datasets}
We chose the SemEval-2014 restaurant review (Rest-14) \cite{pontiki-etal-2014-semeval}, SemEval-2015 restaurant review (Rest-15) \cite{pontiki-etal-2015-semeval}, SemEval-2016 restaurant review (Rest-16) and the multi-aspect multi-sentiment (MAMS) \cite{cheng2017aspect} datasets for sentence-level aspect category and aspect category sentiment. The Rest-14 data has five categories as food, service, ambience, price, and miscellaneous. Rest-15 and Rest-16 have restaurant, ambience, food, service, and drinks categories. MAMS dataset has food, ambience, price, service, miscellaneous, staff, menu, and place categories. The test data size for all the dataset is reported in Table \ref{table-1}. Imbalance signifies the ratio between the largest class size and smallest class size. 

\paragraph{Data for BERT Post-training} 
For BERT post-training, we consider the \textit{Citysearch} data created by, \citet{ganu2009beyond} which contains $\approx$ 2.8 million unlabelled restaurant reviews.

\subsection{Word Representations}
We consider a pre-trained language model called BERT \cite{devlin2019bert} (in particular we chose the `\textit{bert-base-uncased}' model which has 110M parameters). BERT follows a transformer model \cite{vaswani2017attention} for its representation, where the model predicts the masked words using the surrounding context words. We obtain vector representation for each word of a given sentence using BERT language model. We use BERT for both word representations and the post-training tasks. 

\subsection{Baselines} 
We compare the performance of the proposed model with diverse types of baselines such as random, supervised, and weakly supervised methods. 

\begin{itemize}

\item \textbf{Random:} At first, we present a random baseline where the predictions are generated using a uniform distribution. This will provide us with a lower bound for our evaluation. 

\item \textbf{Supervised:} A recent supervised method, \textbf{ACSA-generation} \cite{liu2021solving} solves the ACSA as a generation task. The training and test set are structured with some pre-defined templates. Finally, the authors used BART \cite{lewis2020bart}, a denoising autoencoder, for generating the desired outputs. This will give us an approximate upper bound for our evaluation.  

\item \textbf{Weakly Supervised:} A weakly supervised method, \textbf{JASen} \cite{huang2020weakly} takes unlabelled training reviews and a few keywords corresponding to each aspect categories and sentiment polarity and outputs an (aspect, sentiment) pair for each review. The authors only considered the sentences with single aspect category. 

\item \textbf{Extremely Weakly Supervised:} The method \textbf{X-Class} \cite{wang2021x} takes reviews and a single keyword per class name as inputs and predicts a single class for each review. The method was validated majorly on different topic modelling datasets.
\end{itemize}

\begin{table*}[]
\centering
\resizebox{\textwidth}{!}{%
\begin{tabular}{c|cc|cccc|cccc}
\hline
\multirow{2}{*}{\textbf{}} &
  \multicolumn{1}{c|}{\multirow{2}{*}{\textbf{Supervision Type}}} &
  \multirow{2}{*}{\textbf{Methods}} &
  \multicolumn{4}{c|}{\textbf{ACD}} &
  \multicolumn{4}{c}{\textbf{ACSA}} \\ \cline{4-11} 
 &
  \multicolumn{1}{c|}{} &
   &
  \textbf{Rest-14} &
  \textbf{Rest-15} &
  \textbf{Rest-16} &
  \textbf{MAMS} &
  \textbf{Rest-14} &
  \textbf{Rest-15} &
  \textbf{Rest-16} &
  \textbf{MAMS} \\ \hline
\multirow{4}{*}{\textbf{Baselines}} &
  \multicolumn{2}{c|}{\textbf{Random}} &
  22.50 &
  21.12 &
  19.03 &
  16.45 &
  08.40 &
  08.46 &
  07.16 &
  05.39 \\ \cline{2-3}
 &
  \multicolumn{1}{c|}{\textbf{Supervised}} &
  \textbf{ACSA-Generation} &
  91.41 &
  83.56 &
  87.11 &
  89.23 &
  78.43 &
  71.91 &
  73.76 &
  70.30 \\ \cline{2-3}
 &
  \multicolumn{1}{c|}{\textbf{Weakly Supervised}} &
  \textbf{JASen} &
  42.27 &
  33.29 &
  43.43 &
  21.57 &
  26.62 &
  19.44 &
  23.23 &
  14.74 \\ \cline{2-3}
 &
  \multicolumn{1}{c|}{\begin{tabular}[c]{@{}c@{}}\textbf{Extremely} \\ \textbf{Weakly Supervised}\end{tabular}} &
  \textbf{X-Class} &
  46.69 &
  40.35 &
  36.58 &
  36.52 &
  34.44 &
  25.49 &
  24.83 &
  16.32 \\ \hline
\multirow{4}{*}{\textbf{Proposed}} &
  \multicolumn{1}{c|}{\multirow{4}{*}{\begin{tabular}[c]{@{}c@{}}\textbf{Extremely} \\ \textbf{Weakly Supervised}\end{tabular}}} &
  \textbf{X-SABSA} &
  56.16 &
  58.87 &
  42.77 &
  37.72 &
  
  39.66 &
  42.55 &
  31.46 &
  19.60 \\ \cline{3-3}
 &
  \multicolumn{1}{c|}{} &
  \textbf{AX-SABSA} &
  69.57 &
  56.17 &
  45.69 &
  39.33 &
  
  44.14 &
  40.24 &
  32.23 &
  18.55 \\ \cline{3-3}
 &
  \multicolumn{1}{c|}{} &
  \textbf{X-MABSA} &
  61.73 &
  \textbf{62.07} &
  49.02 &
  56.48 &
  
  44.96 &
  \textbf{44.35} &
  35.81 &
  27.28 \\ \cline{3-3}
 &
  \multicolumn{1}{c|}{} &
  \textbf{AX-MABSA} &
  \textbf{74.90} &
  60.08 &
  \textbf{50.63} &
  \textbf{60.82} &
  
  \textbf{49.68} &
  42.74 &
  \textbf{36.47} &
  \textbf{29.74} \\ \hline
\end{tabular}%
}
\caption{Comparative Results for the ACD and End-to-End ACSA tasks. We report F1-macro score for ACD and F1-PN macro score for ACSA. X-SABSA: Proposed single label predictor model. AX-SABSA: Proposed single label predictor model where the candidate word for each class is also updated. X-MABSA: Proposed multi-label predictor model. AX-MABSA: Proposed multi-label predictor model where the candidate word for each class is also updated. Clustering algorithm used: mini batch k-means for ACD, and gmm for ACSA.}
\label{table-2}
\vspace{-4mm}
\end{table*}

\section{Experimental Evaluation} 
\label{sec:bibtex} 
In this section, we study the performance of the different algorithms on four datasets, compare them with different baselines, and discuss the qualitative analysis of our model performance.

\subsection{Evaluation Framework}
We evaluate our method in an \textbf{End-to-End} framework. The popularly used ABSA evaluation uses gold aspects as a part of input to predict the sentiment polarity of each gold aspect. However, when the task is unsupervised (almost), we do not expect to know the aspect categories beforehand, as has been explored in previous works involving sentiment mining alone. Thus, we follow the End-to-End framework, which has two stages. In the first stage, given sentences, all the aspects are predicted. In the second stage, for each predicted aspect in the first stage, the corresponding sentiment polarity is predicted. Therefore, in our case, the first-stage output is the ACD output, which outputs aspect categories corresponding to each sentence. The second stage output is the ACSA output, which is a set of tuples consisting of (aspect category, sentiment polarity) pairs for each sentence. Therefore, if both the aspect category and sentiment polarity are predicted correctly then only, we consider it as a correct prediction. Thus, the performance is measured over all tuples (aspect, sentiment) in the gold data.

\subsection{Evaluation Metrics}
We consider two metrics for performance evaluation. For the ACD task, we report macro-averaged F1 score (or F1-macro) which is the average of F1-scores per class. For the ACSA task, we report macro-averaged F1-PN score (or macro F1-PN) which is the mean of F1-scores of all aspect category, sentiment (positive, negative) pair tuples. The macro F1-PN is commonly used in different SemEval tasks \cite{pontiki-etal-2016-semeval}.  

\subsection{Empirical Results}  
Comparative results of the ACD and ACSA tasks on different datasets are presented in Table \ref{table-2}. The results show that we achieve far better performance than random baselines given that our approach is unsupervised. The improvement of our multi-label models (X-MABSA and AX-MABSA) is statistically significant at p < 0.01 using paired t-test \cite{hsu2014paired} compared to proposed single label models (X-SABSA and AX-SABSA) and weakly supervised baselines (X-Class, and JASen).

For the ACD task, we achieve baseline results for all the datasets (ACSA module). We obtain F1-macro of 46.69, 40.35, 36.58, and 36.52 on Rest-14, Rest-15, Rest-16, and MAMS dataset, respectively. The proposed X-SABSA model improves the performance significantly on all the datasets (F1-macro of 56.16, 58.87, 42.77, and 37.72 on Rest-14, Rest-15, Rest-16, and MAMS data, respectively). Within our proposed models, we find that our multi-label model X-MABSA performs better than single-label model X-SABSA on all datasets. Especially, on the MAMS dataset, it improves the performance significantly (F1-macro of 56.48). We also observe that the AX-MABSA model (i.e., when automatically selected candidate words are considered for class representation) further improves performance on Rest-14, Rest-16, and MAMS datasets (F1-macro of 74.90, 50.63, and 60.82). It shows that the AX-MABSA model is more generalized and works very well when class names are not present in the input data. 

As the ACSA task is framed as an End-to-End pipeline, we expect the performance to be lower than the often-used ACSA evaluation procedure. We achieve the baseline results (ACSA module) which are F1-PN-macro scores of 34.44, 25.49, 24.83, and 16.32 on Rest-14, Rest-15, Rest-16, and MAMS, respectively. We find that the proposed X-SABSA model improves the performance significantly over the baseline (F1-PN-macro of 39.66, 42.55, and 31.46 on Rest-14, Rest-15, and Rest-16, respectively). The multi-label model, X-MABSA improves the results further (F1-PN-macro of 44.96, 44.35, 35.81, and 27.28, respectively). We also observe that the AX-MABSA model improves the performance on Rest-14, Rest-16, and MAMS data (F1-PN-macro of 49.68, 36.47, and 29.74, respectively).  

We observe that our proposed model performs significantly better than the random, and two weakly supervised baselines (X-Class and JASen) on both ACD and ACSA tasks. As our method is an extremely weakly supervised method, we do not expect our model to be better than the supervised model. However, in comparison to the supervised model (ACSA-generation), our method shows promising performance. For example, on the Rest-14 data, the supervised model achieves an F1-macro of 91.42 while our proposed model achieves an F1-macro of 74.90 for the ACD task. For the ACSA task, the proposed method performs decently compared to the supervised baseline. For example, on Rest-15, the supervised method achieves an F1-PN-macro of 71.91 while our method achieves an F1-PN-macro of 44.35. 

\begin{table*}[]
\centering
\small
\begin{tabular}{p{9.0cm}p{3cm}p{2.6cm}}
\hline
\textbf{Review} & \textbf{Actual} & \textbf{Predicted}\\
\hline\hline
The sashimi is always fresh and the rolls are innovative and delicious. & (food, positive) & (food, positive)\\\hline
While there’s a decent menu, it shouldn’t take ten minutes to get your drinks and 45 for a dessert pizza. & (food, positive),\newline (service, negative) & (food, positive),\newline (service, positive)\\\hline
Who can’t decide on a single dish,
the tapas menu allowed me to
express my true culinary self. & (food, negative),\newline
(menu, positive) & (menu, negative)\\\hline
Roof: very nice space (although I know 5 other rooftop bars just as good), but the crowd was bunch of posers and the owner was a tool. & (place, positive),\newline
(miscellaneous, neutral) & (place, positive),\newline
(ambience, negative)\\\hline
Endless fun, awesome music, great staff! & (service, positive),\newline
(ambience, positive),\newline
(restaurant, positive) & (service, positive),\newline
(ambience, positive)\\\hline
\end{tabular}
\caption{Illustration of the proposed method using few examples}
\label{table-4}
\end{table*}

It is evident that our proposed method works comparatively poorly for ACSA task on the MAMS data. The reason for this is the presence of a remarkably high number of `neutral' classes (43.62\% of total polarity labels). Selecting a single representative surface word for `neutral' class is difficult as there is no association between any word and neutral sentences as compared to the `positive' and `negative' class. For example, the word `bad' can be a representative of `negative' class and the word `good' can be the same of `positive' class, but we found no such representative word for neutral class to perform well. 

\subsection{Performance Analysis}
We report a few example texts with original and our model predicted tags in Table \ref{table-4}. We find that in some cases our model combines two closely related categories to one. For example, the text ``\textit{who can’t decide on a single dish, the tapas menu allowed me to express my true culinary self.}" has gold category as \textit{food} and \textit{menu}. Our model predicts it as \textit{menu}. The reason is that both the words `dish', and `menu' got higher similarity score to category `menu' which is reasonable. 

The fourth sentence in the Table \ref{table-4} has gold labels as `(place, positive)' and `(miscellaneous, neutral)'. Our model predicts as `(place, positive)' and `(ambience, negative)'. We see here that the `miscellaneous' class has been miss-classified into `ambience' and `neutral' to `negative'. The `miscellaneous' class is difficult to represent, even if we replace this word by the automatic surface word selection algorithm. Also, from the sentence, we can sense that the `ambience' can be a class with `negative' polarity. 

The fifth sentence in Table \ref{table-4} has gold labels as `service', `ambience' and `restaurant'. However, our model predicts it as `service', and `ambience' missing the `restaurant' category. This happened as in the sentence, there is no explicit presence of restaurant related words. Another point is that there are some mutual words related to both `ambience' and `restaurant', such as the word `place' can be related to both `ambience' and `restaurant'.

\section{Conclusion}
In this paper, we studied extremely weakly supervised aspect category sentiment analysis across four benchmark datasets, and presented the state-of-the-art unsupervised framework without the requirement of any labelled data. Our method relied only on the surface text of aspect class names and unlabelled texts to extract aspect-sentiment pairs via a multi-label generator model. We proposed an automatic class-representative surface word selection algorithm to select proper representative words corresponding to each class. We also found that unsupervised post-training of language models on domain-specific data improved the word-representations and thus improved the performance. Experiments show that our proposed method performs better than all weakly supervised baseline models. In the future, we intend to improve our methods to incorporate more sentiment classes. We believe that our work would foster more research interest towards unsupervised ABSA.

\section{Limitations}
The main limitation of the proposed work is that it is unable to model the ``neutral'' sentiment class, and performs significantly lower when the number of neutral sentiment reviews are high in a dataset. This is evident from ACSA results on the MAMS data (in Table \ref{table-2}), where the number of neutral classes is high. We have also tried with some possible neutral class related seed category words like `okay', `moderate', `average', etc. but the performance did not improve. It shows that these words can not represent the `neutral' class. Thus, modelling the `neutral' class efficiently will improve the model performance. Although our model performs better than other weakly supervised baselines, there is enough scope for improvement to bridge the gap between the supervised methodologies.

\bibliography{emnlp2022}
\bibliographystyle{acl_natbib}

\end{document}